\documentclass[letterpaper, 10 pt, conference]{ieeeconf}  

\IEEEoverridecommandlockouts                              
\usepackage{geometry}

\usepackage{graphicx} 
\usepackage{epsfig} 
\usepackage{amsopn}
\usepackage{bm} 
\usepackage{amsmath} 
\usepackage{amssymb}  
\usepackage{subcaption}
\usepackage{float}
\usepackage{algorithm}
\usepackage{algpseudocode}
\usepackage{dblfloatfix}
\usepackage{gensymb}
\usepackage{array}
\usepackage[dvipsnames]{xcolor}
\usepackage{multirow}
\graphicspath{{images/}{../images/}}
\usepackage[
    style=ieee,
    sorting=none
]{biblatex}
\newcolumntype{P}[1]{>{\centering\arraybackslash}p{#1}}
\newcolumntype{M}[1]{>{\centering\arraybackslash}m{#1}}
\newtheorem{definition}{Definition}
\newgeometry{left = 54pt, top = 54pt, bottom = 54pt , right =54pt}
\title{ \bf
Fast Staircase Detection and Estimation using 3D Point Clouds with Multi-detection Merging for Heterogeneous Robots}
\author{Prasanna Sriganesh$^{1}$, 
Namya Bagree$^{2}$, 
Bhaskar Vundurthy$^{1}$, 
 and  
Matthew Travers$^{1}$
\thanks{ $^{1}$Prasanna Sriganesh, Bhaskar Vundurthy and Matthew Travers are from The Robotics Institute, Carnegie Mellon University, USA. \texttt{\{pkettava, pvundurt, mtravers\}@andrew.cmu.edu}}
\thanks{ $^{2}$Namya Bagree is from the Department of Mechanical Engineering, Carnegie Mellon University, USA \texttt{nbagree@andrew.cmu.edu}}
}
\begin{document}
\maketitle

\begin{abstract}

Robotic systems need advanced mobility capabilities to operate in complex, three-dimensional environments designed for human use, e.g., multi-level buildings.  Incorporating some level of autonomy enables robots to operate robustly, reliably, and efficiently in such complex environments, e.g., automatically ``returning home'' if communication between an operator and robot is lost during deployment. This work presents a novel method that enables mobile robots to robustly operate in multi-level environments by making it possible to autonomously locate and climb a range of different staircases. We present results wherein a wheeled robot works together with a quadrupedal system to quickly detect different staircases and reliably climb them. The performance of this novel staircase detection algorithm that is able to run on the heterogeneous platforms is compared to the current state-of-the-art detection algorithm.  We show that our approach significantly increases the accuracy and speed at which detections occur. 

                    
\end{abstract}

\begin{keywords}
Robot Perception, Staircase Detection, Point Cloud Estimation
\end{keywords}


\section{Introduction} 




Mobile autonomy and advanced mobility are active research areas within the broader robotics community. However, they have primarily been treated as separate research fields. Approaching this as a tightly-coupled system will help augment the capabilities of mobile robot systems to operate in complex, three-dimensional environments designed for humans.

In the 2020 DARPA Subterranean Challenge, teams of heterogeneous robots, comprising wheeled, legged, and aerial robots, explored unknown urban spaces. Challenges like these typically operate in complex environments composed of underground spaces and multi-level buildings. Operator-less environments require the robots to perceive, analyze and make decisions autonomously. Environment-aware autonomy allows for coordinating robot actions by communicating information about traversable terrains like troughs, inclines, and staircases. Navigating staircases is a significant component of handling multi-floor environments. Knowledge of staircases from wheeled robots that scout enables legged robots to navigate staircases and explore environments that are not reachable by the scout robots. Achieving such real-time coordination with advanced mobile robots requires a fast, robust method of staircase detection.

% ...

\begin{figure}[t]
    \centering
    \begin{subfigure}[t]{.41\linewidth}
        \centering
        \includegraphics[width = 0.99\linewidth]{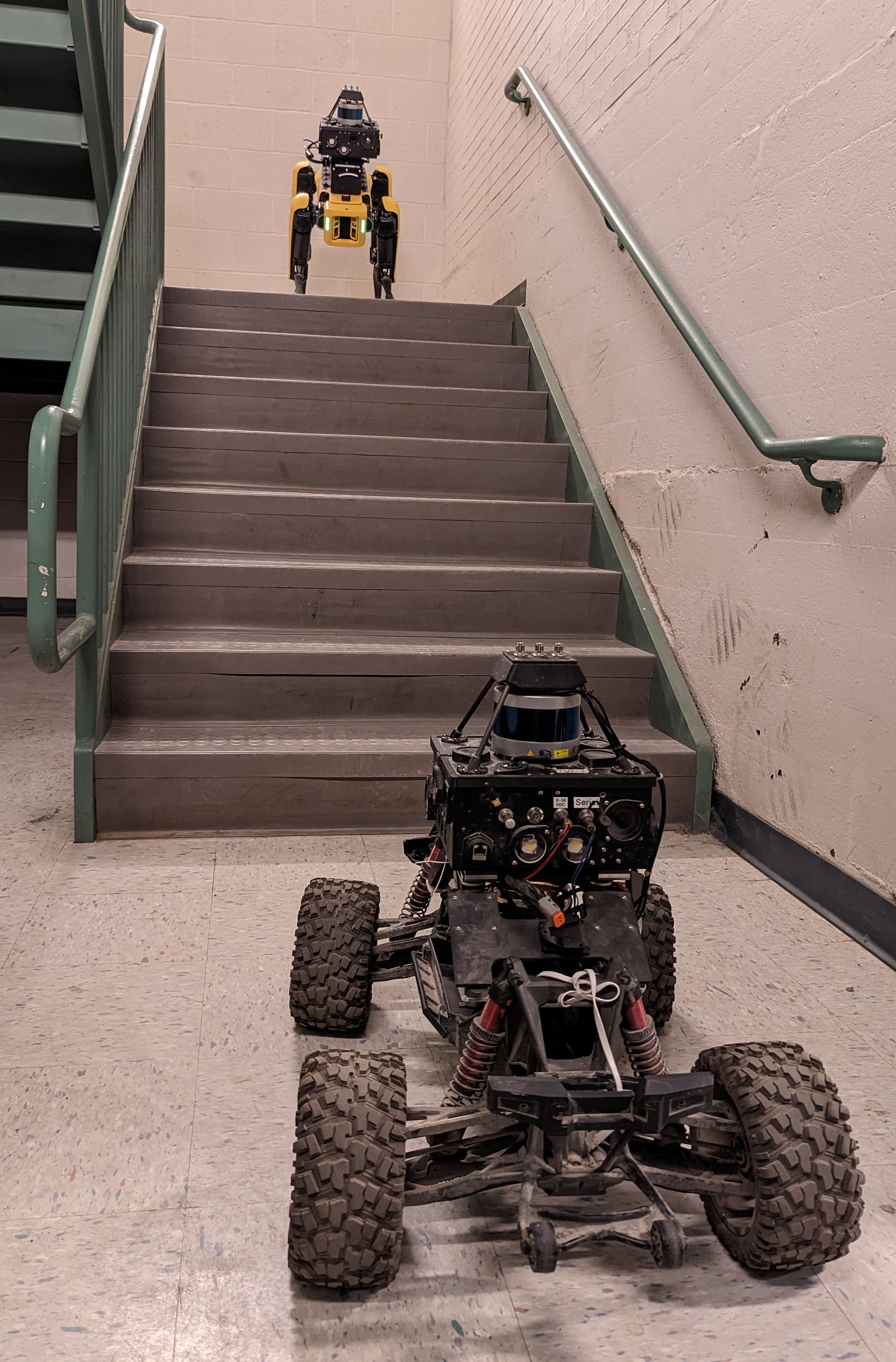}
        \caption{Two heterogeneous robots around a staircase}
    \end{subfigure}
    \begin{subfigure}[t]{.504\linewidth}
        \centering
         \includegraphics[width=0.99\linewidth]{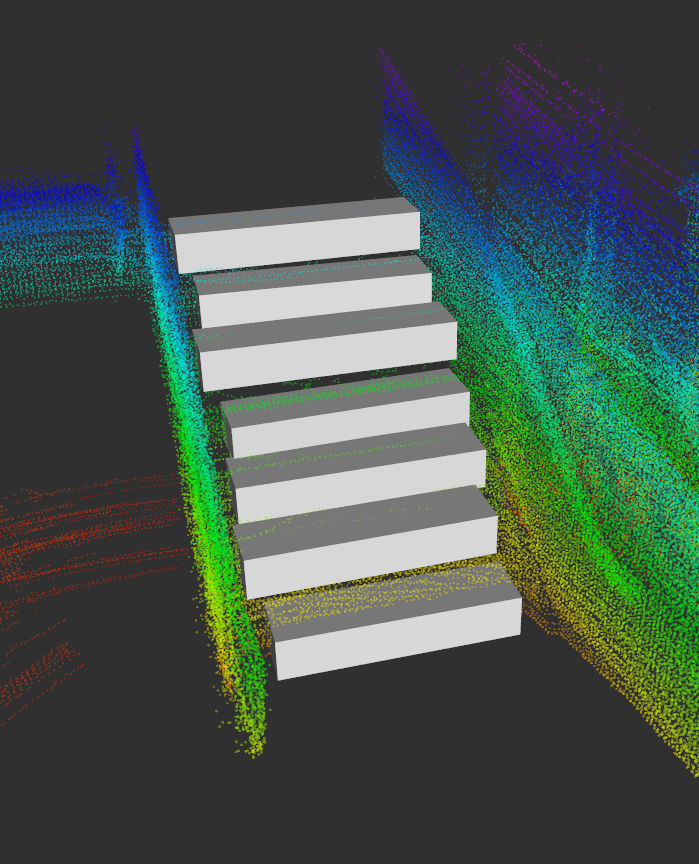}
        \caption{White marker shows the detected staircase}
    \end{subfigure}
    \vspace{0.2cm}
    \caption{Staircase environment and detection}
    \label{fig:mainResult}
    \vspace{-1.0em}
\end{figure}

Staircase detection and characterization are particularly challenging for autonomous robot perception systems. Diversity in staircases, for instance, spiral or hollow staircases, aggravates autonomous detection even further. Existing approaches have emphasized accuracy with a compromise on the speed of detection. Specifically, these approaches can be too slow when implemented in compute-constrained mobile robots. In this work, we focus on detecting a variety of staircases as fast as possible on mobile robots. Further, we aim to accurately indicate the number of steps in a staircase and the size of the steps to help assess if the staircase can be traversed.
\begin{figure*}[!t]
    \centering
    \begin{subfigure}[t]{.2\linewidth}
        \centering
        \includegraphics[width=3.4cm]{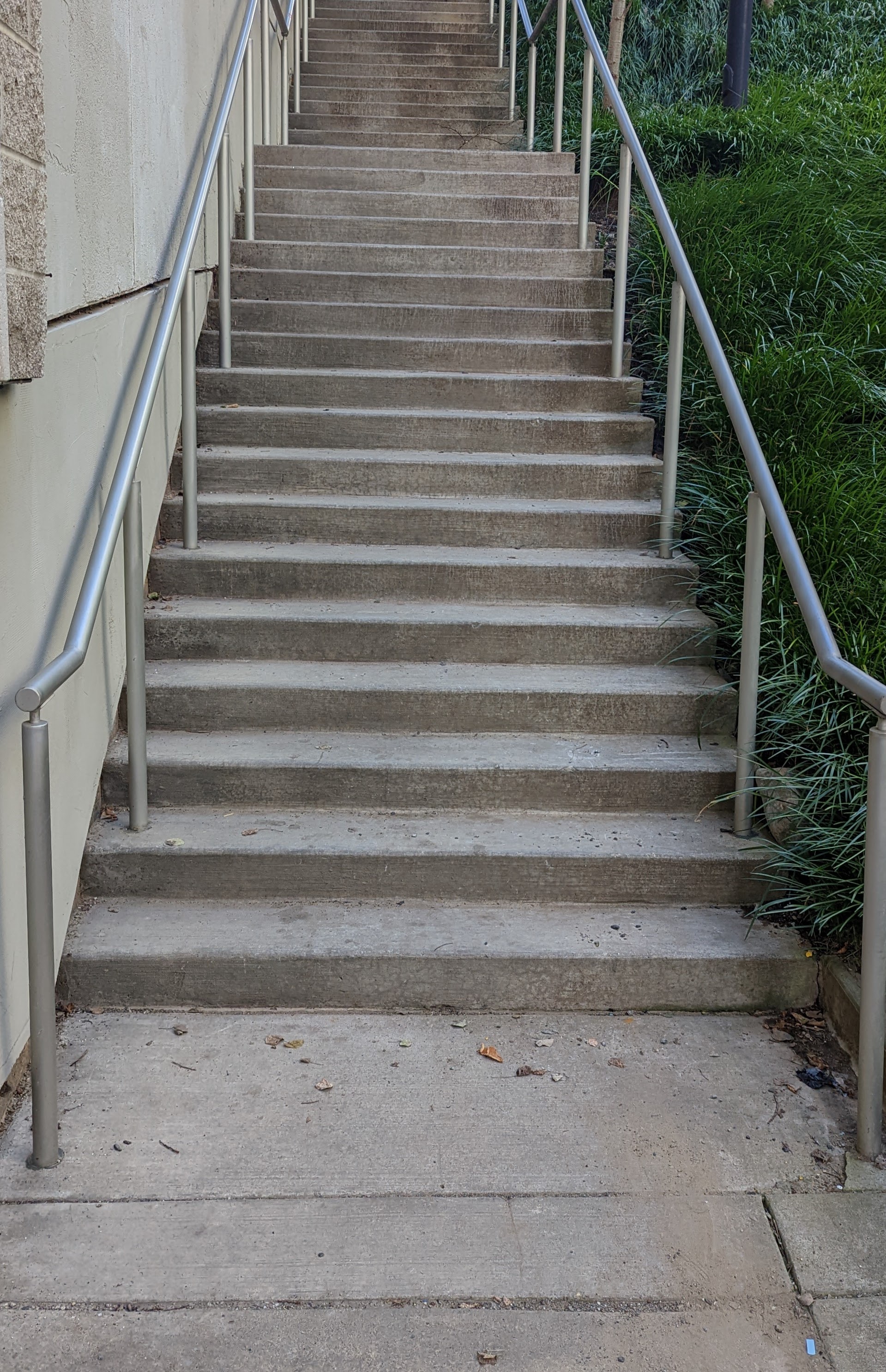}
        \caption{Ascending Staircase}
    \end{subfigure}
    \begin{subfigure}[t]{.19\linewidth}
        \centering
         \includegraphics[width=2.95cm]{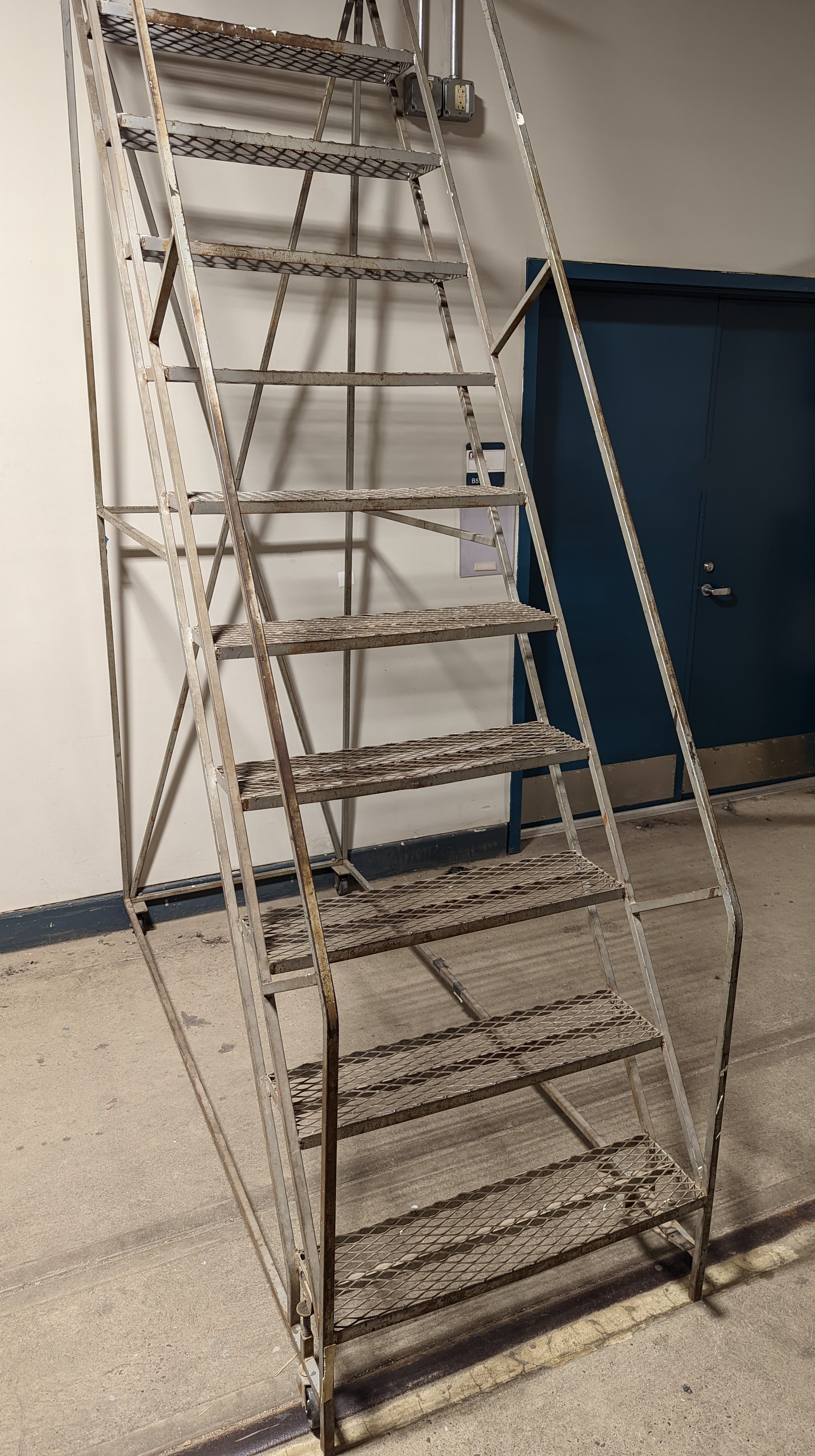}
        \caption{Hollow Staircase}
    \end{subfigure}
    \begin{subfigure}[t]{.19\linewidth}
        \centering
         \includegraphics[width=3.28cm]{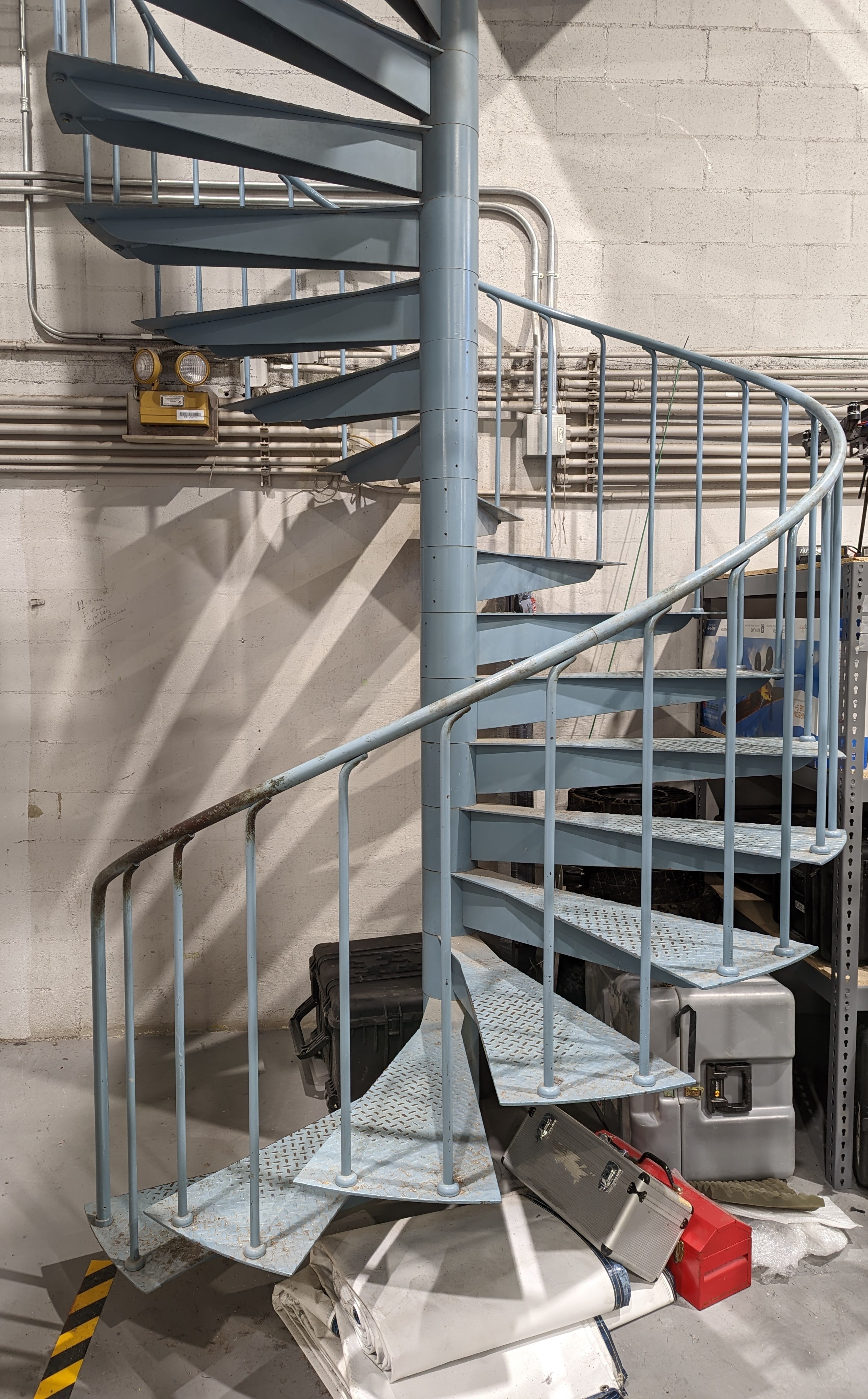}
        \caption{Spiral Staircase}
    \end{subfigure}
    \begin{subfigure}[t]{.19\linewidth}
        \centering
         \includegraphics[width=3.26cm]{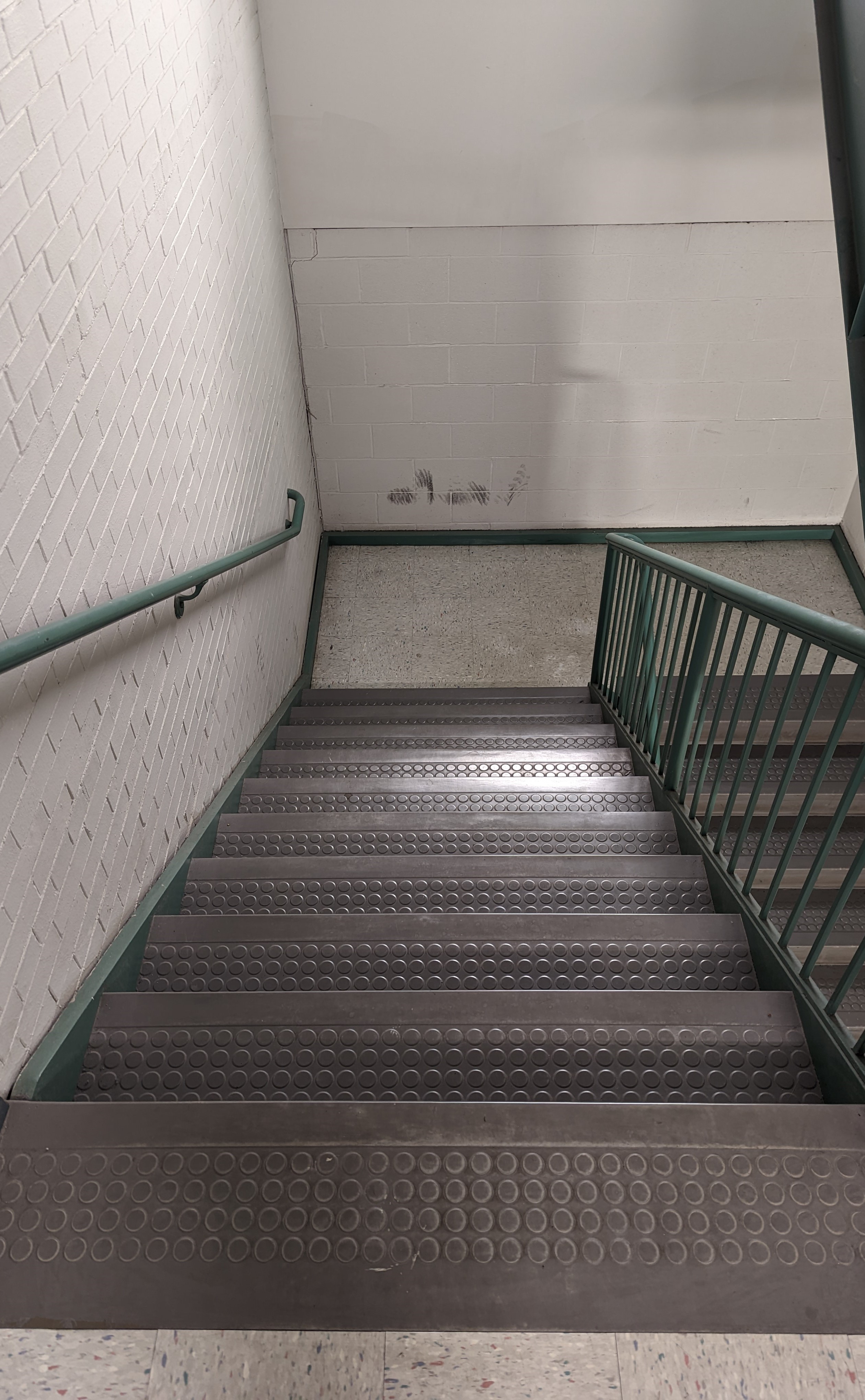}
        \caption{Descending Staircase}
    \end{subfigure}
     \begin{subfigure}[t]{.19\linewidth}
        \centering
         \includegraphics[width=3.18cm]{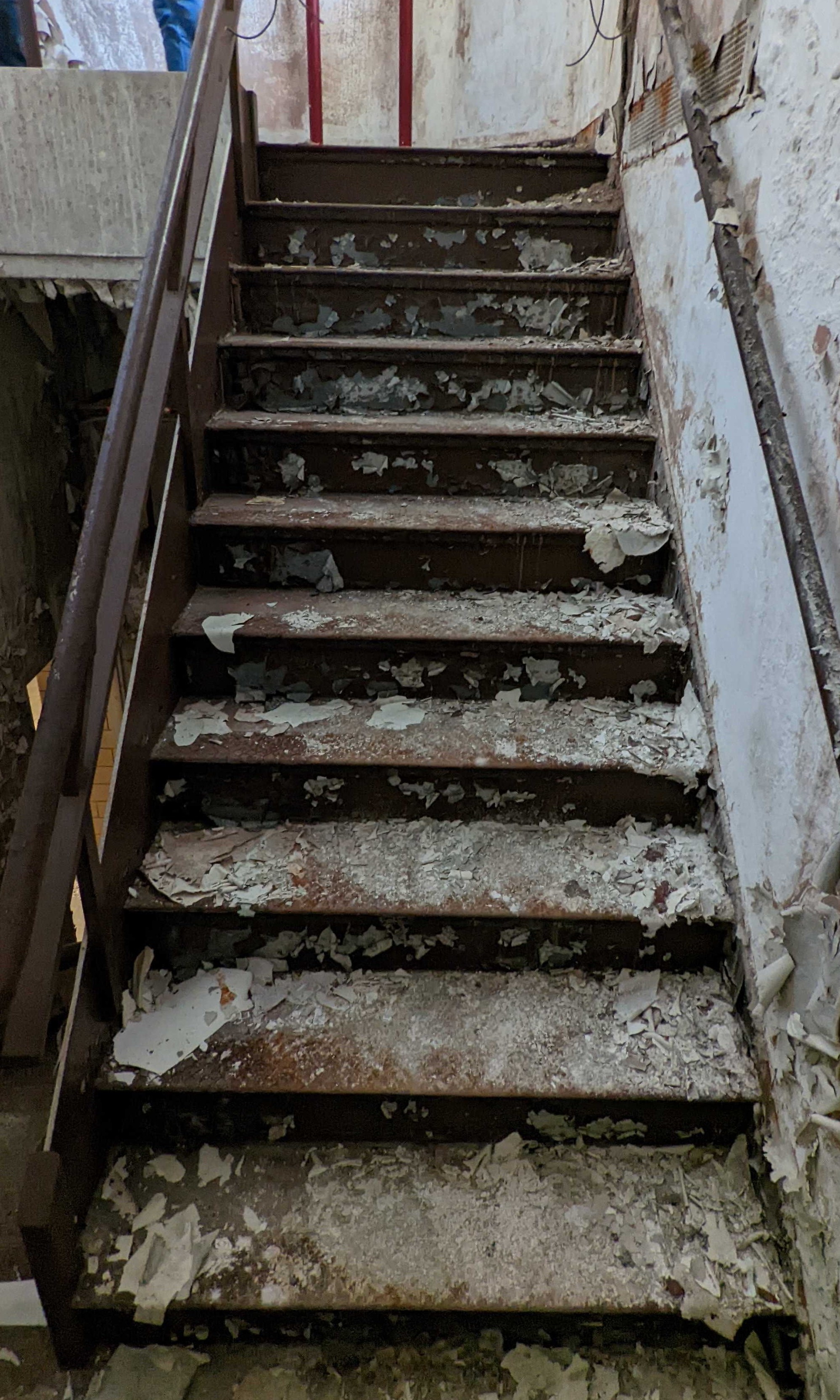}
        \caption{Staircase with Debris}
    \end{subfigure}
    \vspace{0.2cm}
    \caption{Different types of staircases found in an urban environment including industrial and damaged staircases}
    \label{fig:stairs}
\end{figure*}

The main contributions of this work are as follows:
\begin{itemize}
    \item An algorithm to detect different types of staircases in real-time and estimate their geometry.
     \item An algorithm that combines two detection instances of the same staircase obtained by different robots or different viewpoints.
\end{itemize}

\section{Related Work} \label{RW}

 There has been considerable research carried out in the field of staircase detection. All the methods either use images or point clouds as their primary modality. Once the staircases are detected, the next task is estimating the features of all detected stairs. While detection is fairly indifferent between images and point clouds, estimation benefits immensely from the latter.

 One of the first image based approaches to staircase detection was put forth by Cong et al. \cite{cong2008stairway} where the authors detect the edges formed by stairways in image space \cite{zhong2011stairway}\cite{harms2015detection}\cite{samakming2008development}. 
 Murakami et al. \cite{Murakami} used both RGB images and depth images to segment edges and detect both ascending and descending staircases. Even contemporary learning based algorithms have been used to detect staircases from images \cite{patil}\cite{Ilyas}.

The only advantage of image based methods is the speed of the detections. Some shortcomings include reliance on images, which makes the system environment dependent. In low-light conditions, the robustness of the system goes down. Learning methods require massive datasets to work reliably. Moreover, the lack of depth information in RGB images prohibits accurate geometry or location estimation. The camera positioning also impacts the field of view of detection.

The primary idea of detecting staircases with point clouds is by segmenting planes formed by the stairs as they have a fixed geometry. Point clouds help also estimate the geometry and location of the staircase which can be used as an input for navigation. Point clouds are typically collected from LiDAR sensors and Depth cameras. O\ss wald et al. \cite{osswald20113d} first explored and experimented on two different plane segmentation methods to detect risers of the staircase.

Different variations of Random Sample Consensus (RANSAC) has also been extensively used to segment planes and then detect staircases \cite{vlaminck2013obstacle}\cite{qian2014ncc}. RANSAC based methods have been used to estimate staircase location for navigation by different robots \cite{sharma2019begin}\cite{woo2019stair}\cite{sanchez2021staircase}. Fourre et al. \cite{fourre2020autonomous} even devised a way to localize industrial stairways which have no risers. Even though RANSAC is a simple and efficient algorithm, it is non-deterministic. It does not guarantee a best-solution or have a fixed time-bound. This is not a great way to detect stairways in scenarios that are time critical. These algorithms also have prerequisites on which part of the staircase needs to be visible to the sensor.

Westfechtel et al. \cite{westfechtel2018robust} were the first to successfully achieve detection and estimation of staircases in all direction($360^o$). They compared three different segmentation methods to detect planes in LiDAR point clouds and used a graph-based strategy to detect staircases of all types. Their estimates of the staircase location and the geometry were the best among all the previous work. Consequently, we can consider this work to be the current state-of-the-art for staircase detection. Although, their biggest drawback was the speed of the detection. The robot was expected to be static during the entire process, and plane analysis took around 4-8 seconds. This is entirely not feasible in search-and-rescue scenarios where every second is crucial.

None of the plane-based methods discuss detection speed, which is essential to us. All previous works look at staircase detection as a one-off algorithm. There has not been much research conducted on multiple robot use scenarios. Two robots with different viewpoints can help achieve a better estimation of a staircase by fusing both instances. These are two aspects we intend to address.


\section{Methodology} \label{Me}
In this section, we present an algorithm that detects staircases from LiDAR point clouds and accurately estimates the staircase parameters such as location, height, and depth. The main intuition behind our algorithm is to segment only the edges formed by the staircase surfaces. The pipeline has three major steps: pre-processing, segmentation, and detection.

\begin{figure*}[!t]
    \centering
    \begin{subfigure}[t]{.18\linewidth}
        \centering
        \includegraphics[width=2.93cm]{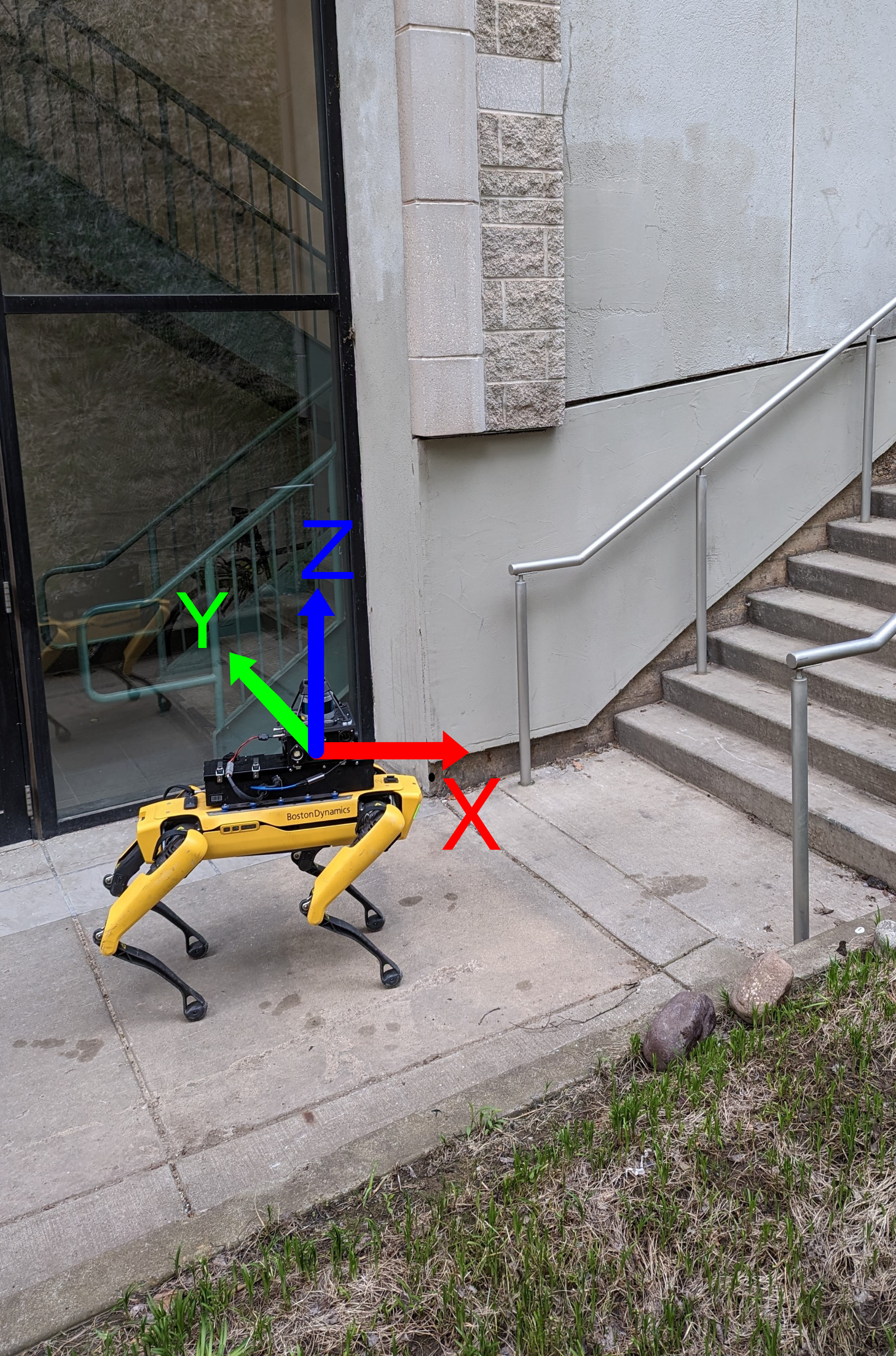}
        \caption{Robot and its reference frame}
        \label{fig:spot_with_axes}
    \end{subfigure}
    \begin{subfigure}[t]{.19\linewidth}
        \centering
         \includegraphics[width=3.22cm]{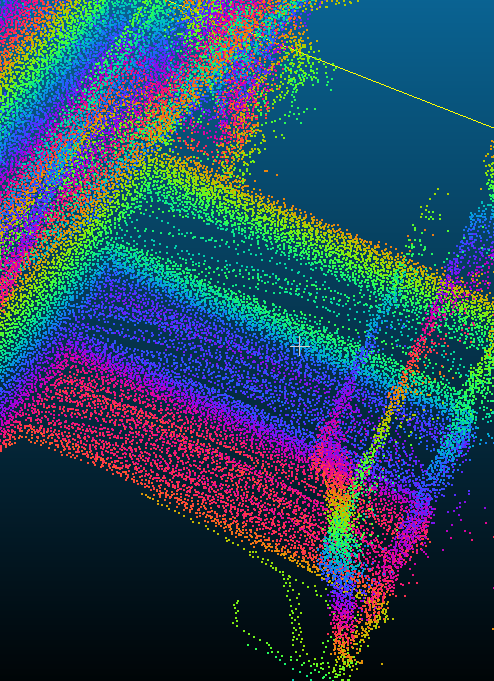}
        \caption{Input point cloud of the staircase}
        \label{fig:input_pc}
    \end{subfigure}
    \begin{subfigure}[t]{.20\linewidth}
        \centering
         \includegraphics[width=3.5cm]{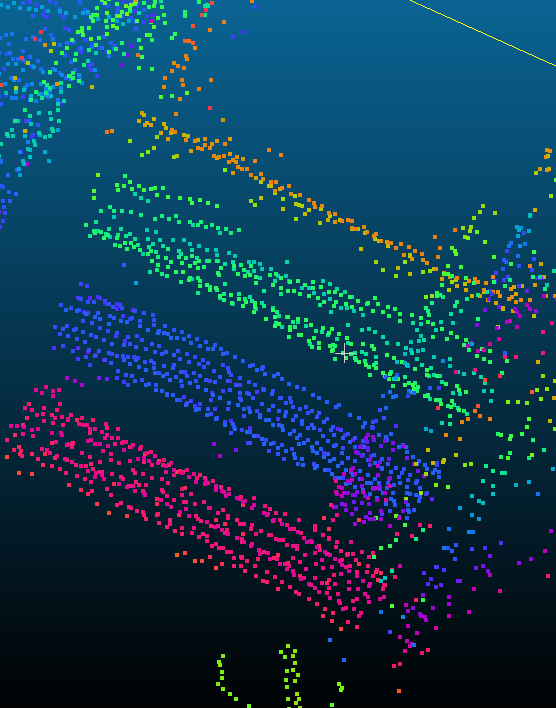}
        \caption{Point cloud after first step of pre-processing}
        \label{fig:pre_process1}
    \end{subfigure}
    \begin{subfigure}[t]{.20\linewidth}
        \centering
         \includegraphics[width=3.36cm]{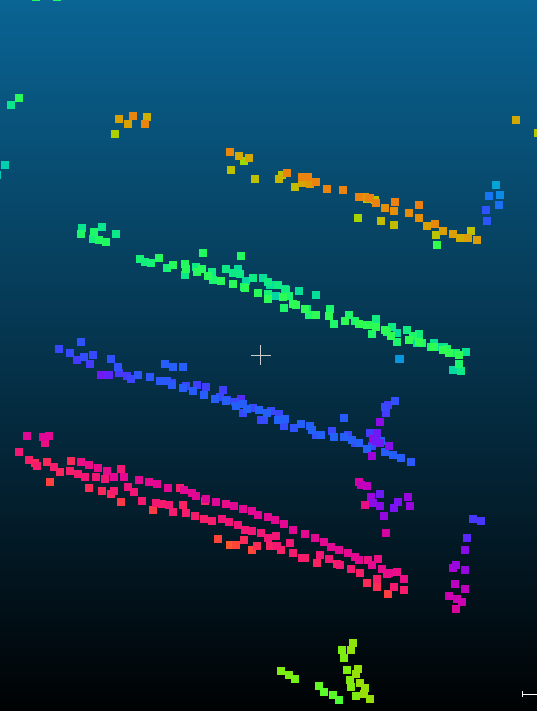}
        \caption{Result point cloud after all pre-processing steps}
        \label{fig:pre_process2}
    \end{subfigure}
     \begin{subfigure}[t]{.19\linewidth}
        \centering
         \includegraphics[width=3.15cm]{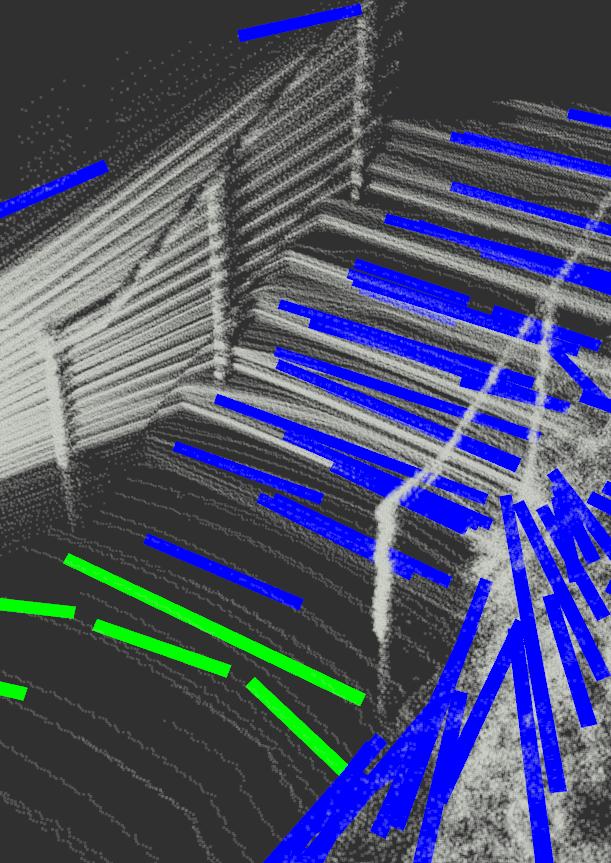}
        \caption{Segmented lines, grouped based on height}
        \label{fig:seg_result}
    \end{subfigure}
    \vspace{0.2cm}
    \caption{Preprocessing and segmentation overview for stair detection}
    \label{fig:ppp}
\end{figure*}

\subsection{Pre-Processing}
Let's first define the robot's frame. As shown in Fig. \ref{fig:spot_with_axes}, the x-axis is in line with the robot orientation, the z-axis faces upwards and y-axis is to its left. The input point cloud (accumulated from multiple LiDAR scans) is converted into a 3D voxel grid with fixed leaf size. This input cloud is shown in Fig. \ref{fig:input_pc}.

The first step in pre-processing is to retain points with maximum height for a given X-Y cell index. Specifically, for each column in the Z direction, we retain only the one with the highest z-value. This step retains all the points visible from a top view of the point cloud. All the planes perpendicular to the ground are reduced to a line parallel to the ground.
We apply this technique to the point cloud shown in Fig. \ref{fig:input_pc}. The resulting point cloud is shown in Fig. \ref{fig:pre_process1}.
 

Next we organise this point cloud into a 2D array around the robot using cylindrical coordinates. The z-axis is discretised into rows of the array, while the columns are indexed using the azimuth angle $\theta$ which is given by $tan^{-1}(\frac{y}{x})$. If there are 2 or more points with similar $z$ and $\theta$, we then compute the range $\rho ~(= \sqrt{x^2 + y^2})$ of the point and retain the point with smallest range value. This essentially reduces all the horizontal planes (stairs) into a single edge. 
 
The two pre-processing steps should reduce the 3D point cloud to points that belong to the stair edges. Fig. \ref{fig:pre_process2} shows the final processed cloud. 
 
\subsection{Segmentation}

As mentioned before, the point cloud is organized into a 2D array structure that can be thought of as an unwrapped cylinder with the robot at its center. Each row index corresponds to a fixed $z$ height, and the column indices represent the azimuth angle spanning from $-180\degree$ to $180\degree$. Consequently, a full row of this array can be treated as single 2D laser scan. Furthermore, since all points in each row have approximately equal z-values, we can fit lines in the $xy$ space and add the height information later.

To extract lines from a 2D laser-scan, we use a modified version of Iterative-End-Point Fit \cite{ramer1972iterative}\cite{linereview}. Given $N$ points in a scan, we fit a line between the first and last points to start the splitting. We then find the point with the maximum distance to this line. If the distance is more than a threshold ($d_p$), the points are split into two groups until the number of points in each set is greater than a limit ($N_{min}$) or if all the points are at a distance less than a threshold $d_p$.

Typically, if all the points are close to the line, the loop stops and returns the line. We modify this behavior to use weighted line fitting to estimate a line with all the points. Weighted line fitting is a version of least square line fitting with the uncertainty of a point used as weights. We refer to \cite{weightedline} for a detailed discussion of the problem.

The merging is usually done if the points forming the two lines are collinear. We modify this to use the merging criteria provided by weighted line fitting. The main advantage of weighted line fitting is that it outputs a covariance for each line. This allows us to estimate the similarity between two lines better and merge it better. We represent each line fit using this method by the parameters below:
\begin{itemize}
    \item 3D start point of the line - $\Bar{p_s}$
    \item 3D end point of the line - $\Bar{p_e}$
    \item Orientation $\alpha$ - angle between the line with the $xy$ plane
    \item Covariance Matrix $P_L$
\end{itemize}

This line segmentation algorithm is run separately on each row (2D scan) of the point cloud array. All the resulting lines are added to a single list $\mathcal{L}$. As a result, we get lines parallel to the ground plane. Since staircase edges also have to be parallel to the ground, it eliminates the need to check for line altitudes. If the point cloud has $Z$ rows and $T$ columns, the time complexity of this algorithm is $O(ZTlog(T))$.

After all the lines are segmented, we group them into three groups based on their height in the 3D space. Assuming that the robot has a fixed height, it is trivial to group lines that are above the ground ($\mathcal{L}_{ag}$), below the ground ($\mathcal{L}_{bg}$), and that are part of the ground plane ($\mathcal{L}_{g}$). This makes it easier to search for staircases that are ascending and descending. Fig. \ref{fig:seg_result} shows the lines detected by our algorithm in that scene. Blue represents lines above the ground, and the green represents lines on the ground plane.

\subsection{Detection}

\begin{figure}[!t]
    \centering
    \begin{subfigure}[t]{.5\linewidth}
        \centering
        \includegraphics[width=0.99\linewidth]{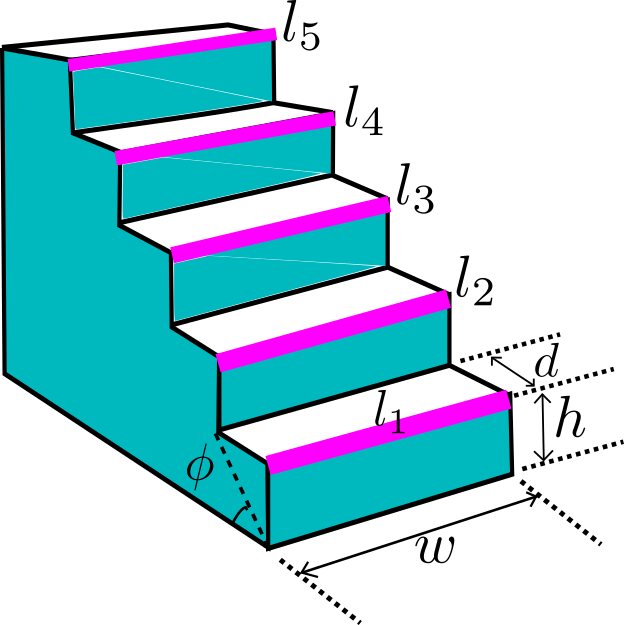}
        \caption{Staircase model with parameters, pink lines represent stair edges to be segmented}
        \label{f:stair_param}
    \end{subfigure}
    \begin{subfigure}[t]{.48\linewidth}
        \centering
         \includegraphics[width=0.9\linewidth]{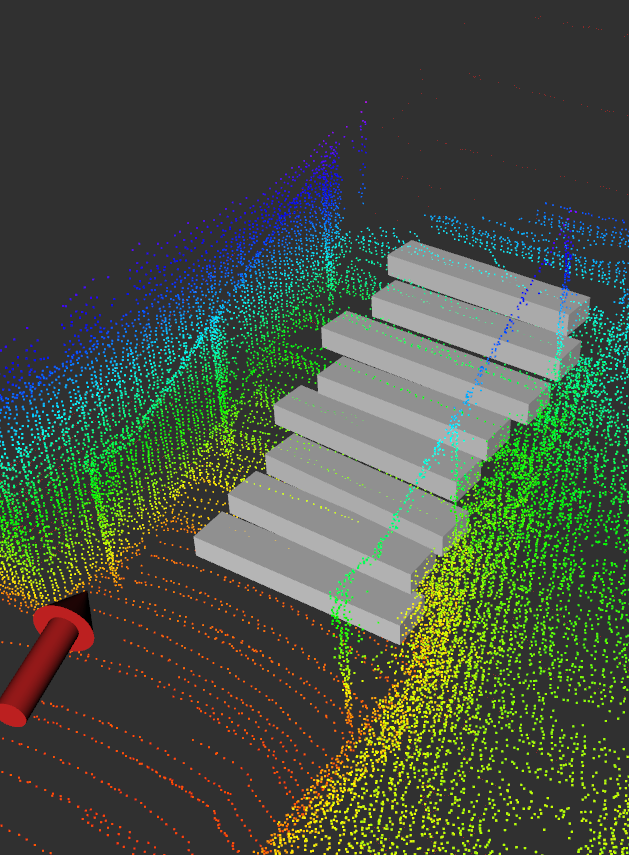}
        \caption{Detected staircase, each white marker corresponds to a stair}
        \label{fig:det_stair}
    \end{subfigure}
    \vspace{0.2cm}
    \caption{Staircase model and successful detection result}
\end{figure}

We start searching for staircases in the segmented lines from the previous step. We first describe the model of our staircase as follows. Fig. \ref{f:stair_param} also describes these parameters.
\begin{itemize}
    \item Step Height, $h$
    \item Step Depth, $d$
    \item Step Width, $w$
    \item List of Lines $\mathcal{L}$, Each line $l_i$ represents a stair and is described by 3D start ($\Bar{p_s}^{(i)}$) point, 3D end ($\Bar{p_e}^{(i)}$) point and it's orientation ($\alpha^{(i)}$) in the XY plane
    \item Stair Slope, $\phi = tan^{-1}(h/d)$
\end{itemize}

 In order to start grouping a set of lines that form a staircase, we define a stair case based on the standards set by OSHA\cite{osha}. 
 \begin{definition}
 \label{def:staircase}
 Given two lines $l_1 (\Bar{p_s}^{(1)}, \Bar{p_e}^{(1)}, \alpha^{(1)})$ and $l_2 (\Bar{p_s}^{(2)}, \Bar{p_e}^{(2)}, \alpha^{(2)})$, we compute the height $h_i$ given by the difference in $z$ between the lines, the depth $d_i$ given by the xy distance between the lines and the slope $\phi_i = tan^{-1}(h_i/d_i)$. Any two lines that satisfy the five conditions is defined as a \textbf{stair}:
 \begin{enumerate}
    \item $0.11$ m $\leq h_i \leq 0.30$ m 
    \item $0.15$ m $\leq d_i \leq 0.45$ m  
    \item $25 \degree \leq \phi_i \leq 60 \degree$ 
    \item $ |\alpha^{(1)} - \alpha^{(2)}| \leq 10 \degree $
    \item There is no other line in between $l_1$ and $l_2$ \hfill $\blacksquare$
\end{enumerate}
\end{definition}


Depending on whether we want to detect an ascending or descending staircase, we pick the appropriate list ($\mathcal{L}_{ag}$ or $\mathcal{L}_{bg}$) of segmented lines. The algorithm to detect staircases has two parts, initialization, and extension. In the initialization stage, we create a subset of lines that are below $2.5~h_{max}$ in the z-direction called the initialization list. Here $h_{max}$ is the maximum step height allowable ($0.3~m$). We then look for two lines in this initial list that can form a staircase as per Definition \ref{def:staircase}. If two such lines exist, we add them to a staircase list $\mathcal{S}$.

Once we have the first two steps of the staircase, we perform an extension. For every line in the increasing z-direction, we check if it forms a staircase with the previous line in the staircase set using Definition \ref{def:staircase}. If it complies, we add the new line to the set and repeat the extension until all lines are exhausted. If there are more than four lines in the staircase list, it is a successful staircase detection, and use this to estimate the staircase geometry.

If the initialized staircase does not extend, i.e., the set only has two stairs, we remove the lines from the initialization list and repeat the process until the initialization list is empty. The pseudo-code of this method is presented in \textbf{Algorithm \ref{alg:stair}}. To detect descending staircases, we switch the set of lines to $L_{bg}$ and run the same algorithm, but the lines are checked in decreasing z-direction. Fig. \ref{fig:det_stair} shows the detected staircase.

 \begin{algorithm}[!t]
    \caption{Detect ascending staircase from list of lines}\label{alg:stair}
    \begin{algorithmic}[1]
        \Require{Set of above ground lines - $\mathcal{L}_{ag}$}
        \Ensure{Staircase $\mathcal{S}$ with $N$ lines }
        \State Initialize list $\mathcal{L}_I$ with all lines ($\mathcal{L}_{ag} \leq 2.5h_{max}$)
        \While{$\mathcal{L}_I$ not empty}
         \State Initialize empty staircase set $\mathcal{S}$
            \State $stair_{init}\gets false$
            \While{$stair_{init}$ is $false$}
                \State Pick 2 lines, $l_1, l_2$ from $\mathcal{L}_I$
                \If{$l_1$ and $l_2$ form a stair (Definition \ref{def:staircase})}
                    \State $stair_{init} = true$
                    \State Add $l_1$ and $l_2$ to S
                \EndIf
                \If{No valid pair exits}
                    \State \Return $\mathcal{S}$
                \EndIf
            \EndWhile
            \State Reorder lines in $\mathcal{L}_{ag}$ by ascending order of height(z)
            \For{Every line $l_{curr}$ in $\mathcal{L}_{ag}$}
                \State $\l_{prev} \gets$ last line in $\mathcal{S}$ 
                \If{$l_{curr}$ and $l_{prev}$ form a stair (Definition \ref{def:staircase}) }
                    \State Add $l_{curr}$ to Set $\mathcal{S}$
                \EndIf
            \EndFor
            \If{Total stairs in $\mathcal{S} \geq 4 $}
                \State \Return $\mathcal{S}$
            \Else
                \State Remove the initialized stair lines from $\mathcal{L}_I$
            \EndIf
            \State Initialize empty staircase list $\mathcal{S}$
        \EndWhile
        \State \Return $\mathcal{S}$
    \end{algorithmic}
\end{algorithm}

\subsection{Estimation}

After detection, we estimate the model parameters as shown in Fig. \ref{f:stair_param}. Our detection gives us a list of lines $\mathcal{S}$ with $k$ lines. Assuming that all stairs in a stairway have similar dimensions, we compute the parameters as shown below:
\begin{align*}
    h &= \frac{\sum_{n=1}^{k-1} ||\Bar{p_s}^{(i+1)} - \Bar{p_s}^{(i)}||_z +  ||\Bar{p_e}^{(i+1)} - \Bar{p_e}^{(i)}||_z}{2k} \\
    d &= \frac{\sum_{n=1}^{k-1} ||\Bar{p_s}^{(i+1)} - \Bar{p_s}^{(i)}||_{xy} + ||\Bar{p_e}^{(i+1)} - \Bar{p_e}^{(i)}||_{xy}}{2k} \\
    w &= \frac{\sum_{n=1}^{k}||\Bar{p_s}^{(i)} - \Bar{p_e}^{(i)}||}{k} \\
     & \textit{where } ||p_1 - p_2||_{z} \textit{ is the 1D distance in z axis} \\
     & \textit{where } ||p_1 - p_2||_{xy} \textit{ is the euclidean distance in xy plane}
\end{align*}

We can also define the location of the staircase ($\Bar{S_p}$) at the center of the first stair, given by:
\begin{align*}
     \Bar{S_p} = (\Bar{p_s}^{(1)} + \Bar{p_e}^{(1)}) /2
\end{align*}

\subsection{Multi-Detection Merging}
This section describes a simple algorithm that can merge two different detection instances of the same staircase. The detections can be either by the same robot during different time instances or by using different robots with different viewpoints. We would first like to define a criteria to classify stairs as similar. 
\begin{definition}
 \label{def:line}
 Given two stairs $l_1 (\Bar{p_s}^{(1)}, \Bar{p_e}^{(1)}, \alpha^{(1)})$ and $l_2 (\Bar{p_s}^{(2)}, \Bar{p_e}^{(2)}, \alpha^{(2)})$, we first compute the height $h_i$ given by the difference in $z$ between the stairs, the depth $d_i$ given by the xy distance between the stairs. Any two stairs that satisfy the following three conditions are considered to be the \textbf{same stair}:
 \begin{enumerate}
    \item $h_i \leq 0.05$ m 
    \item $d_i \leq 0.05$ m  
    \item $ |\alpha^{(1)} - \alpha^{(2)}| \leq 10 \degree $ \hfill $\blacksquare$
\end{enumerate}
\end{definition}

\textbf{Algorithm \ref{alg:merge}} describes the way to combine two detection instances. The main idea is to find the location of the intersection for two detections. We do this by iteratively finding two lines, one from each detection, that are similar. This common line (stair) between detections will act as an anchor point to merge the two detections. Once merged, if any more stairs do not have a corresponding pair, it is appropriately appended to the end/start of the list. The staircase parameters are re-estimated after this merge.

 \begin{algorithm}[t!]
    \caption{Merging staircases}\label{alg:merge}
    \begin{algorithmic}[1]
        \Require{Two Staircase Detections $\mathcal{S}_a$ with $k$ stairs and $\mathcal{S}_b$ with $m$ stairs with $k \leq m$}
        \Ensure{Fused Staircase $\mathcal{S}$ if successful, else NO\_MATCH}
        \State $match \gets false$
        \For{Every stair $l_a$ in $\mathcal{S}_{a}$}
            \For{Every stair $l_b$ in $\mathcal{S}_{b}$}
                \If{$l_{a}$ and $l_{b}$ are similar stairs (Definition \ref{def:line}) }
                    \State $match \gets true$ \Comment{Stair match found}
                    \State $i \gets index(l_a)$ and $j \gets index(l_b)$
                    \State \textbf{break} and \textbf{goto} 10
                \EndIf
            \EndFor
        \EndFor
        \State \textbf{if} {$match$ is $false$} \textbf{return} NO\_MATCH 
        \State $\mathcal{S} \gets \{ l_1,\cdots ,merge(l_{i},l_{j}), merge(l_{i+1}, l_{j+1}), \cdots, l_m\}$ where 2 stairs are merged using \cite{weightedline}
        \State Re-estimate parameters of Staircase $\mathcal{S}$ and \textbf{return} $\mathcal{S}$
    \end{algorithmic}
\end{algorithm}

This algorithm allows the merging of any two detections irrespective of the time of the detection or the robot's viewpoint as long as the point clouds are registered in the same global frame, and there is at least one common stair between the detections. 

\section{Experimentation and Results} \label{ER}
\subsection{Experimentation}

\begin{figure}[t!]
    \centering
    \begin{subfigure}[t]{0.35\linewidth}
        \centering
        \includegraphics[trim={5cm 15cm 3cm 5cm},clip, width=2.6cm]{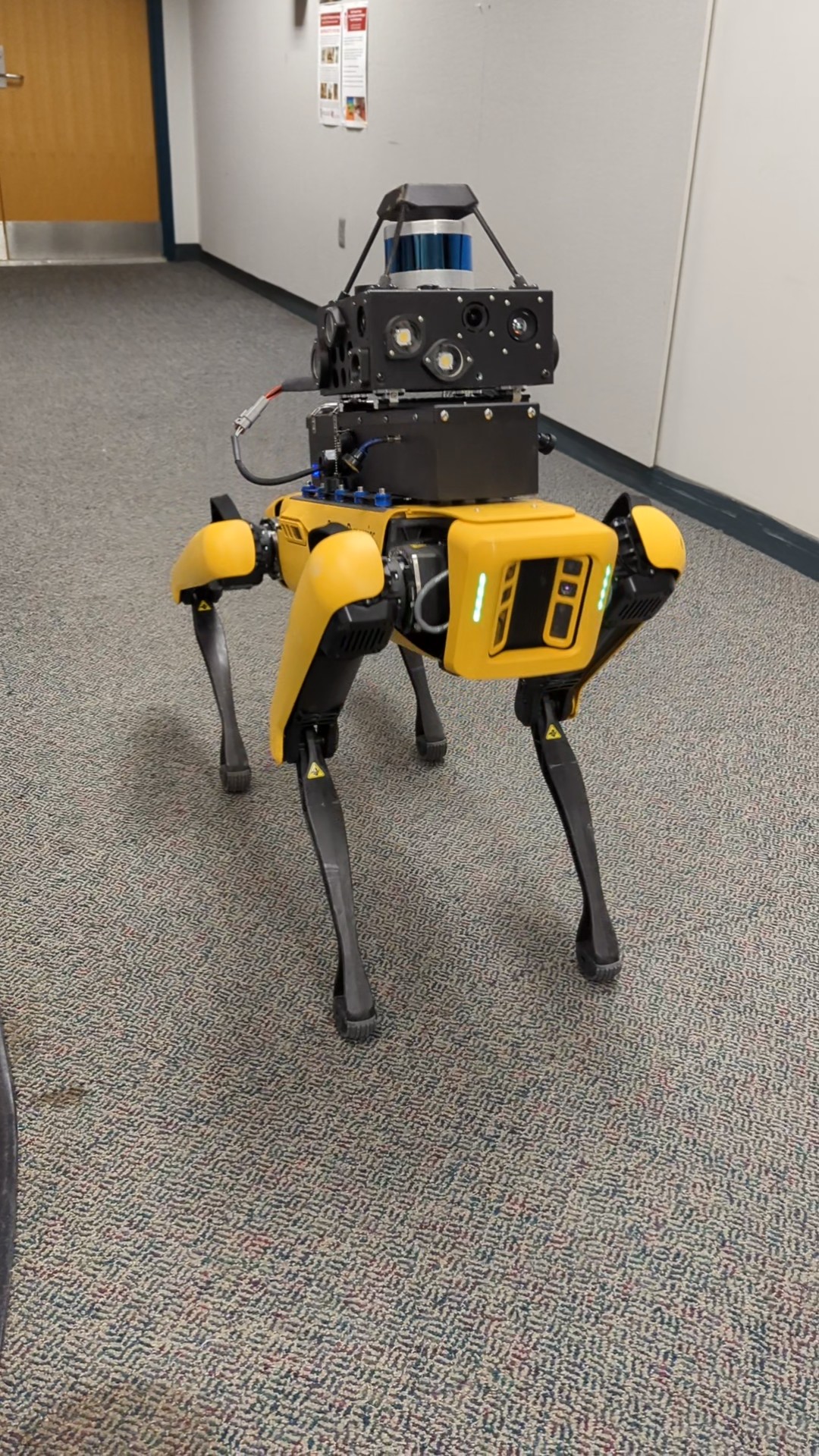}
        \caption{Spot Legged Robot}
        \label{Spot}
    \end{subfigure}
    \begin{subfigure}[t]{0.35\linewidth}
        \centering
       \includegraphics[width = 3.17cm]{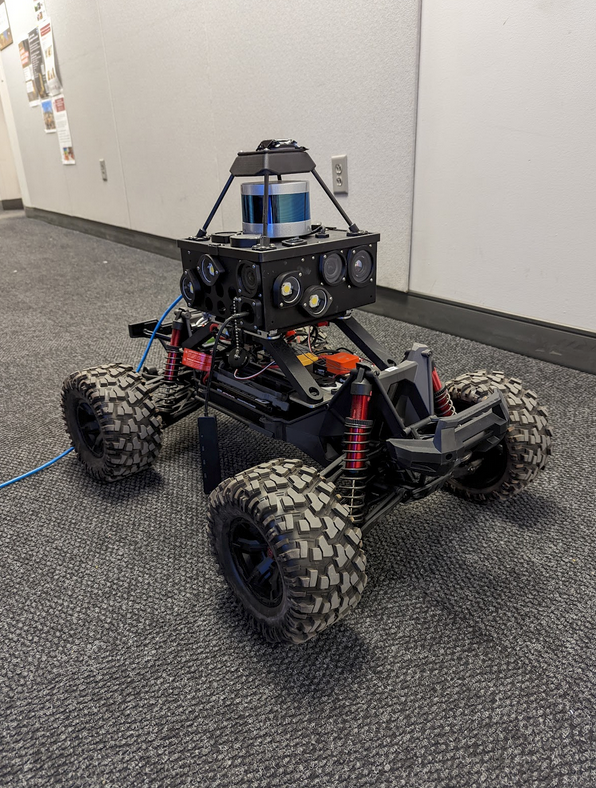}
        \caption{Autonomous Ground Vehicle}
        \label{Traxxas}
    \end{subfigure}
    \vspace{0.2cm}
    \caption{Mobile robot platforms used to test our algorithm}
\end{figure} 

\subsubsection{System Overview}

We make use of heterogeneous robots with identical perception sensor payloads. Fig. \ref{Spot} shows the Boston Dynamics Spot legged robot. This robot is capable of climbing staircases once its location has been estimated. Fig. \ref{Traxxas} shows the unmanned ground vehicle. The vehicle is capable of moving at up to 6m/s autonomously. The sensor payload has four fish-eye cameras on each side and a LIDAR sensor to obtain 3-dimensional scans of the environment. It also houses a Jetson AGX Xavier as the processor that performs SLAM using Super Odometry \cite{zhao2021super} and other autonomy tasks.

\subsubsection{Setup}
We tested our algorithm's performance on staircases shown in Fig. \ref{fig:stairs} and performed a comparison with the state-of-the-art \cite{westfechtel2018robust}.
The leaf size of the voxel grid was set to 2.5 cm. As the robots move around these five staircase types, we create a dataset that spans multiple distance points and orientations in front of the staircase.
We aim to detect all five kinds of staircases, followed by estimating the geometric parameters of the detected stairs. An essential aspect of our comparison is the time taken to obtain a detection.

\begin{figure*}[!t]
    \centering
    \begin{subfigure}[t]{.19\linewidth}
        \centering
        \includegraphics[width=3.3cm]{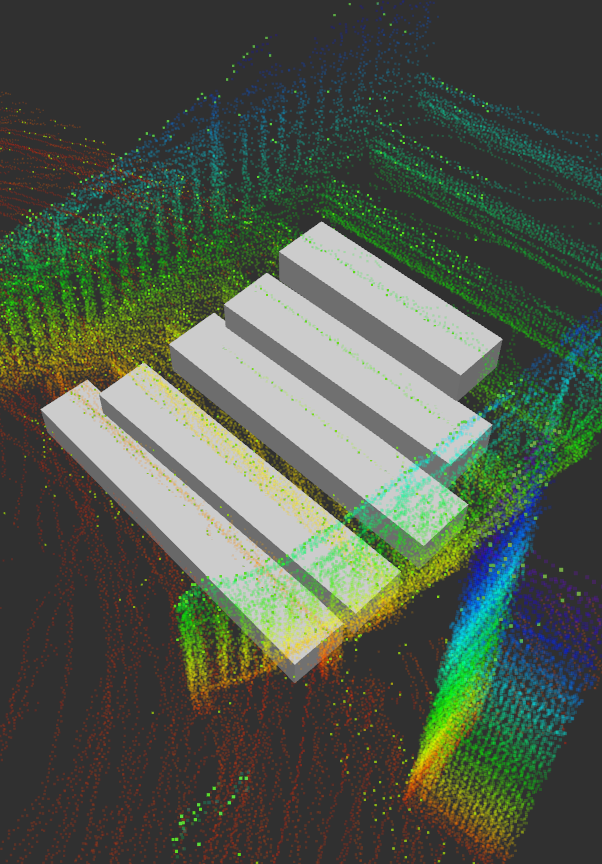}
        \caption{Ascending Staircase}
    \end{subfigure}
    \begin{subfigure}[t]{.18\linewidth}
        \centering
         \includegraphics[width=3.17cm]{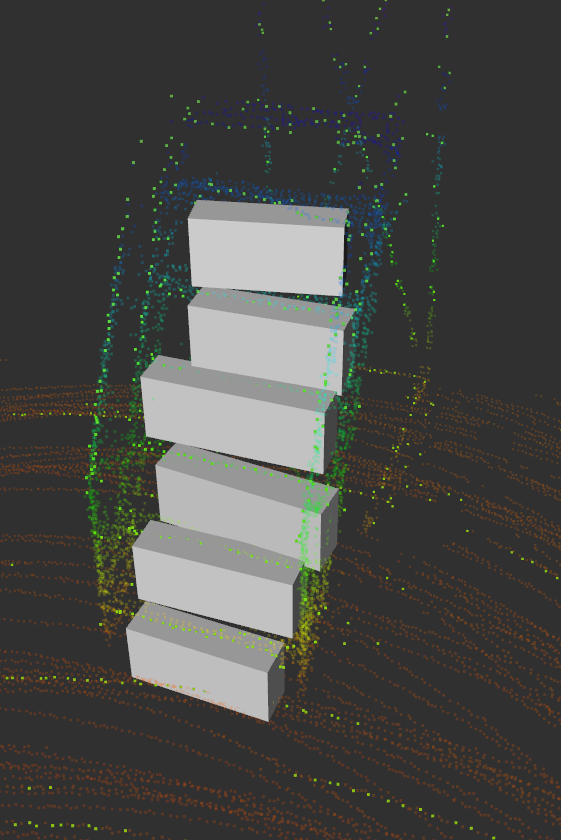}
        \caption{Hollow Staircase}
    \end{subfigure}
    \begin{subfigure}[t]{.20\linewidth}
        \centering
         \includegraphics[width=3.3cm]{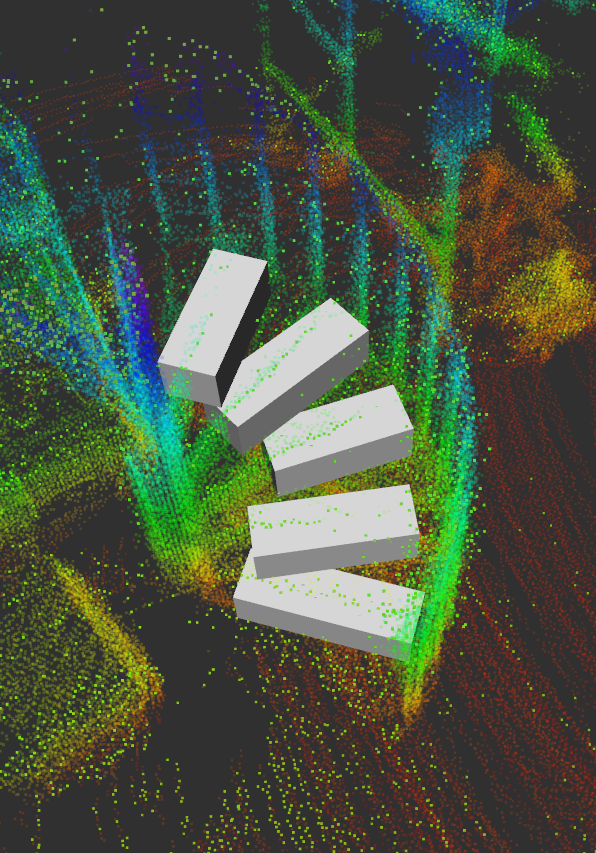}
        \caption{Spiral Staircase}
    \end{subfigure}
    \begin{subfigure}[t]{.19\linewidth}
        \centering
         \includegraphics[width=2.95cm]{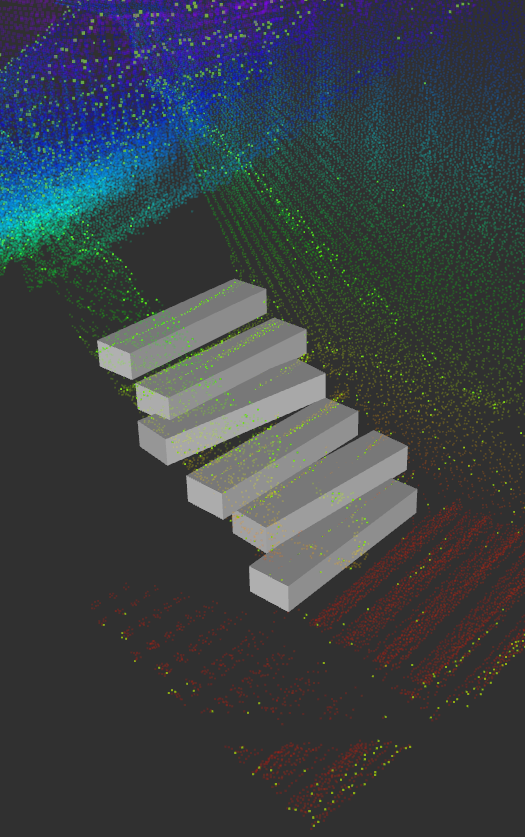}
        \caption{Descending Staircase}
    \end{subfigure}
     \begin{subfigure}[t]{.19\linewidth}
        \centering
         \includegraphics[width=3.4cm]{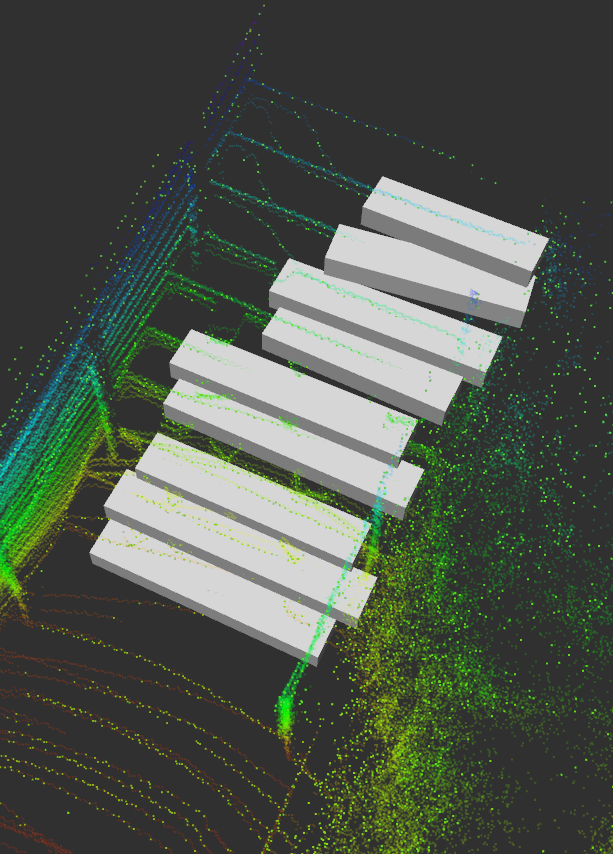}
        \caption{Staircase with Debris}
    \end{subfigure}
    \vspace{0.2cm}
    \caption{Staircase detection results for different types of staircases shown in Fig. \ref{fig:stairs}. }
    \label{fig:detections_all}
\end{figure*}

\begin{table*}[t]
  \centering
  \begin{tabular}{|M{2.5cm}||P{1.4cm}|P{1.3cm}|P{1.3cm}|P{1.3cm}||P{1.4cm}|P{1.3cm}|P{1.3cm}|P{1.3cm}|} \hline
  & \multicolumn{4}{|c||}{\textbf{Our Method}} &  \multicolumn{4}{|c|}{\textbf{Westfechtel et al. \cite{westfechtel2018robust}}} \\ \hline
  \textbf{Staircase Type}  & \textbf{Average Time Taken (ms)} & \textbf{Height Error (cm)} & \textbf{Depth Error (cm)} & \textbf{Width Error (cm)} & \textbf{Average Time Taken (ms)} & \textbf{Height Error (cm)} & \textbf{Depth Error (cm)} & \textbf{Width Error (cm)} \\ \hline
  Ascending Staircase & \textcolor{ForestGreen}{21} & \textcolor{ForestGreen}{1.101} & \textcolor{Red} {2.804} & \textcolor{Red} { 38.125} & \textcolor{Red} {1569} & \textcolor{Red} { 1.527} & \textcolor{ForestGreen}{2.564} & \textcolor{ForestGreen}{7.023} \\ \hline
  Hollow Staircase & \textcolor{ForestGreen}{18} & \textcolor{ForestGreen}{0.982} & \textcolor{ForestGreen}{0.839} & \textcolor{ForestGreen}{15.121} & \textcolor{Red} {1009} & \textcolor{Red} {7.673} & \textcolor{Red} {13.760} & \textcolor{Red} {41.233} \\ \hline
  Spiral Staircase & \textcolor{ForestGreen}{15} & \textcolor{ForestGreen}{0.490} & \textcolor{Red} {3.467} & \textcolor{ForestGreen}{11.040} & \textcolor{Red}{1005} & \textcolor{Red} {3.357} & \textcolor{ForestGreen}{2.687} & \textcolor{Red} {27.667} \\ \hline
  Descending Staircase & \textcolor{ForestGreen}{10} & \textcolor{Red} {2.367} & \textcolor{ForestGreen}{2.653} & \textcolor{Red} {27.773} & \textcolor{Red} {623} & \textcolor{ForestGreen}{1.517} & \textcolor{Red} {4.470} & \textcolor{ForestGreen}{15.437} \\ \hline
  Staircase with Debris & \textcolor{ForestGreen}{48} & \textcolor{ForestGreen}{2.980} & \textcolor{ForestGreen}{2.362} & \textcolor{ForestGreen}{29.612} & \textcolor{Red} {4282} & \textcolor{Red} {3.660} & \textcolor{Red} {3.736} & \textcolor{Red} {93.270} \\ \hline
  \end{tabular}
  \caption{Comparison of parameter estimation errors and time taken by our method vs \cite{westfechtel2018robust} for 5 categories of staircases}
  \label{tab:comparison}
\end{table*}

\begin{table*}[h]
\centering
\begin{tabular}{|c|c|cc|cc|}
\hline
                            &                                    & \multicolumn{2}{c|}{\textbf{Our Method}}                                                                      & \multicolumn{2}{c|}{\textbf{Westfechtel et al.\cite{westfechtel2018robust}} }                                                             \\ \cline{3-6} 
\multirow{-2}{*}{\textbf{}} & \multirow{-2}{*}{\textbf{Samples}} & \multicolumn{1}{c|}{\textbf{Detection}} & \textbf{\begin{tabular}[c]{@{}c@{}}False \\ Positives\end{tabular}} & \multicolumn{1}{c|}{\textbf{Detection}} & \textbf{\begin{tabular}[c]{@{}c@{}}False \\ Positives\end{tabular}} \\ \hline
Ascending Staircase         & 15                                 & \multicolumn{1}{c|}{15}                 & {\color[HTML]{000000} 0}                                            & \multicolumn{1}{c|}{15}                 & {\color[HTML]{000000} 10}                                           \\ \hline
Hollow Staircase            & 17                                 & \multicolumn{1}{c|}{17}                 & {\color[HTML]{000000} 1}                                            & \multicolumn{1}{c|}{3}                 & {\color[HTML]{000000} 1}                                            \\ \hline
Spiral Staircase            & 15                                 & \multicolumn{1}{c|}{15}                 & {\color[HTML]{000000} 0}                                            & \multicolumn{1}{c|}{15}                  & {\color[HTML]{000000} 14}                                            \\ \hline
Descending Staircase        & 14                                 & \multicolumn{1}{c|}{14}                 & {\color[HTML]{000000} 0}                                            & \multicolumn{1}{c|}{12}                 & {\color[HTML]{000000} 3}                                           \\ \hline
Staircase with Debris       & 10                                 & \multicolumn{1}{c|}{10}                 & {\color[HTML]{000000} 1}                                            & \multicolumn{1}{c|}{10}                 & {\color[HTML]{000000} 10}                                           \\ \hline
\end{tabular}
\caption{Performance comparison of our method vs \cite{westfechtel2018robust} for five categories of staircases}
\label{tab:samples}
\end{table*}

\subsection{Results}
Fig. \ref{fig:detections_all} summarizes all the detections from our algorithm in five kinds of staircases. The white markers represent the detected stairs, which are resized to match our algorithm's estimation results. It is worth noting that the algorithm detects all staircases with an average computation time of $20.64~ms$ as opposed to the $1511.33~ms$ using SOTA.

TABLE \ref{tab:comparison} summarizes our algorithm's estimation accuracy and detection speed and compares it to the SOTA. We recorded the average error with respect to the ground truth in estimating height, depth, and width. Green indicates better accuracy for that parameter for the given algorithm. As a general trend, we estimate the height and depth better than the SOTA. While both algorithms trade blows with standard ascending and descending staircases, we perform much better than the SOTA in other cases. Our algorithm has significantly higher estimation accuracy when it comes to hollow or spiral staircases. We also performed relatively better when the staircases had some debris on them. Our detection speeds were faster by at least two orders of magnitude.

\begin{figure}[b!]
    \centering
    \includegraphics[width = 0.65\linewidth]{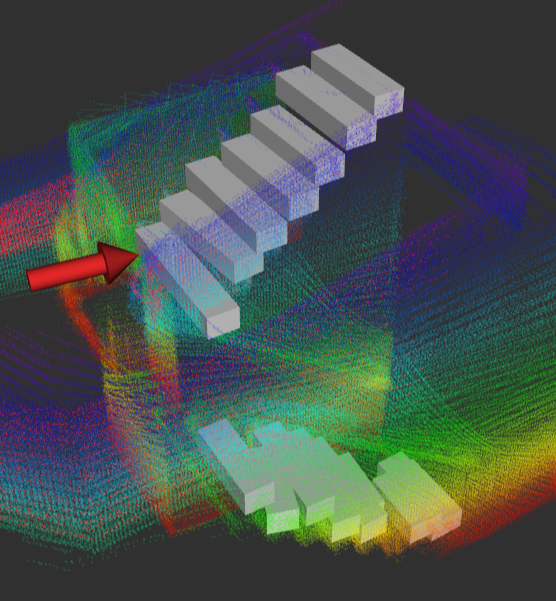}
    \caption{Ascending and descending staircases detected by our algorithm simultaneously}
    \label{fig:as_des_both}
\end{figure}

In TABLE \ref{tab:samples}, we recorded the performance of both algorithms. Both algorithms detect all staircases for standard ascending and descending staircases, but the SOTA has a lot of false positives in the same scene. In the hollow staircase scenario, the SOTA could barely detect it successfully. In addition to this, we were also able to run our algorithm in real-time on the robots. Fig. \ref{fig:as_des_both} shows the result we detected both an ascending and descending staircase simultaneously.

\begin{figure}[b!]
    \centering
    \vspace{-1.0em}
    \includegraphics[width = 0.9\linewidth]{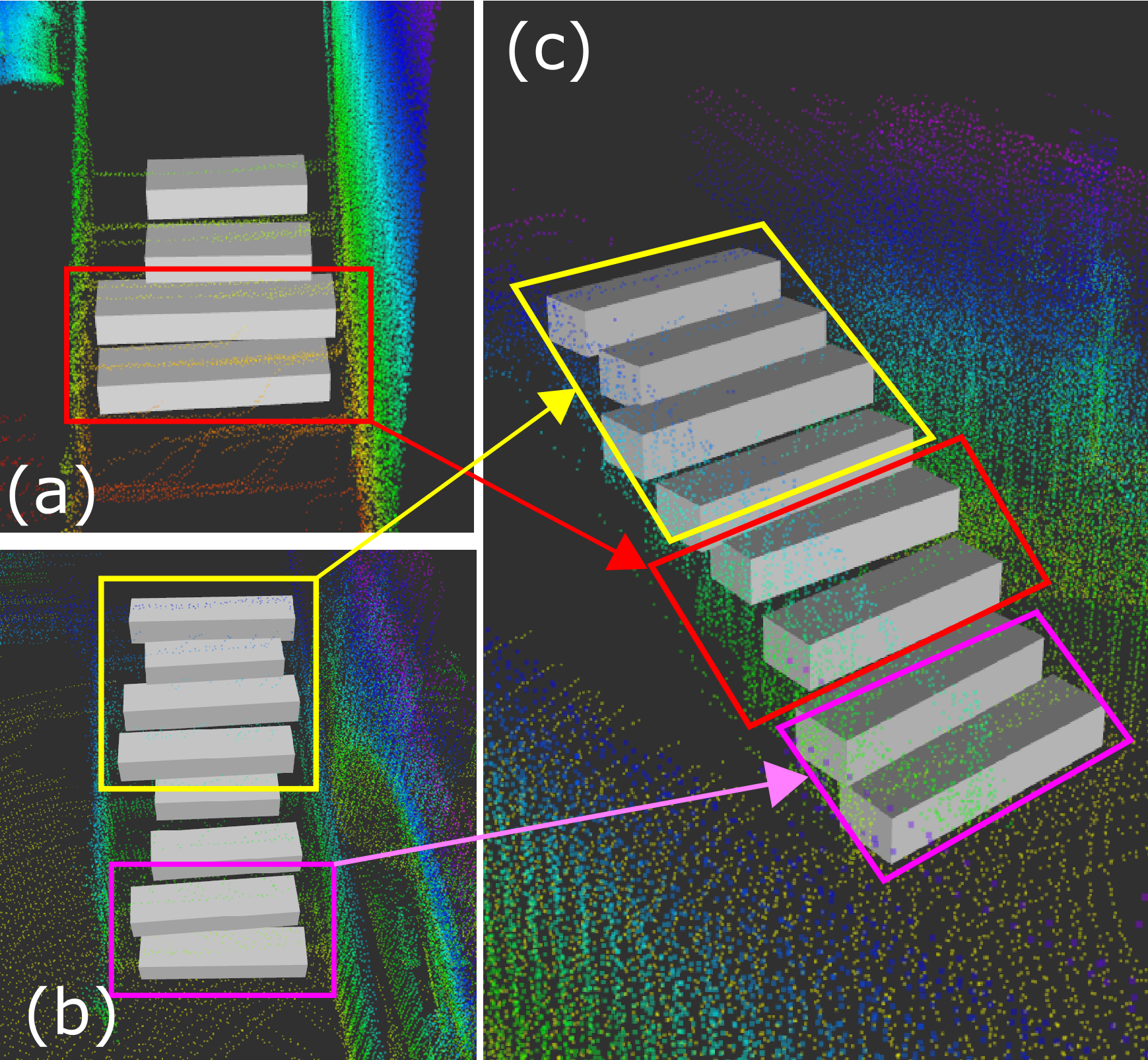}
    \caption{Two detections [(a), (b)] get merged into one good detection (c) using Algorithm \ref{alg:merge}.  $~~~~~~~~~~~~~~~~~~~~~~~~~~~~~~~~~~~~~~~~~~~~~~~$ }
    \label{fig:merge_result}
\end{figure}

We also evaluated the performance of our merging algorithm. We were successfully able to merge two different detection instances into one. Fig. \ref{fig:merge_result} shows the overview of this process. In this case, Fig. \ref{fig:merge_result}(a) was the detection result of the Spot when it was on top of the staircase looking down, and \ref{fig:merge_result}(b) was the output of the ground vehicle from the bottom. We can clearly see a better merged staircase detection in Fig. \ref{fig:merge_result}(c). This algorithm is not only limited to multiple robots but will also work with two different instances of a detection on the same robot.

\section{Conclusions and Future Work} \label{CL}
In this paper, we presented an algorithm to detect and estimate staircases in real-time and merge detections of the same staircase from different robots. We successfully demonstrated our algorithms on two heterogeneous robots and evaluated the accuracy of our estimation. Compared to the state-of-the-art, our algorithm is faster by an order of two magnitudes while maintaining similar accuracy for the staircase parameters. 

One of the limitations of our algorithm is in scenarios where the risers or treads of the stairs are sloped or if the stair edges are curved. In these cases, the segmented lines would be noisy, resulting in failed detections. We want to address these scenarios in future work. Further, we would like to improve the merging algorithm by potentially using a Kalman filter that can help in noisy environments.


\printbibliography
\end{document}